\documentclass{article} %
\usepackage{iclr2025_conference,times}

\usepackage{url}
\usepackage{natbib}
\usepackage[colorlinks=true,citecolor=apolloblue,linkcolor=apolloblue,urlcolor=apolloblue]{hyperref}

\usepackage[utf8]{inputenc} %
\usepackage[T1]{fontenc}    %
\usepackage{booktabs}       %
\usepackage{amsfonts}       %
\usepackage{nicefrac}       %
\usepackage{microtype}      %
\usepackage{xcolor}         %
\usepackage{csquotes}
\usepackage{mathtools}
\usepackage{graphicx}
\usepackage{amsmath,amsfonts,amsthm,bm}

\title{You can remove GPT2's LayerNorm by fine-tuning}

\author{%
Stefan Heimersheim\thanks{\texttt{stefan@apolloresearch.ai}} 
\AND 
\textmd{Apollo Research}
}

\iclrfinalcopy %

\begin{document}

\maketitle

\begin{abstract}
  The LayerNorm (LN) layer in GPT-style transformer models has long been a hindrance to mechanistic
  interpretability. LN is a crucial component required to stabilize the training of large language
  models, and LN or the similar RMSNorm have been used in practically all large language models
  based on the transformer architecture.
  The non-linear nature of the LN layers is a hindrance for mechanistic interpretability as it
  hinders interpretation of the residual stream, and makes it difficult to decompose the model into
  circuits. Some researchers have gone so far as to name \enquote{reasons interpretability researchers
  hate layer norm}.

  In this paper we show that it is possible to remove the LN layers from a pre-trained GPT2-small
  model by fine-tuning on a fraction (500M tokens) of the training data. We demonstrate that this
  LN-free model achieves similar performance to the original model on the OpenWebText and ThePile
  datasets (-0.05 cross-entropy loss), and the Hellaswag benchmark (-0.5\% accuracy). We provide
  our implementation at \href{https://github.com/ApolloResearch/gpt2_noLN}{https://github.com/ApolloResearch/gpt2\_noLN}, and fine-tuned GPT2-small models at
  \href{https://huggingface.co/apollo-research/gpt2_noLN}{https://huggingface.co/apollo-research/gpt2\_noLN}.
  
  Our work not only provides a simplified model for mechanistic interpretability research, but also
  provides evidence that the LN layers, at inference time, do not play a crucial role in transformer
  models.
\end{abstract}

\begin{figure}[hb!]
  \centering
  \includegraphics[width=0.8\textwidth,trim=5cm 1.5cm 7cm 5.5cm,clip]{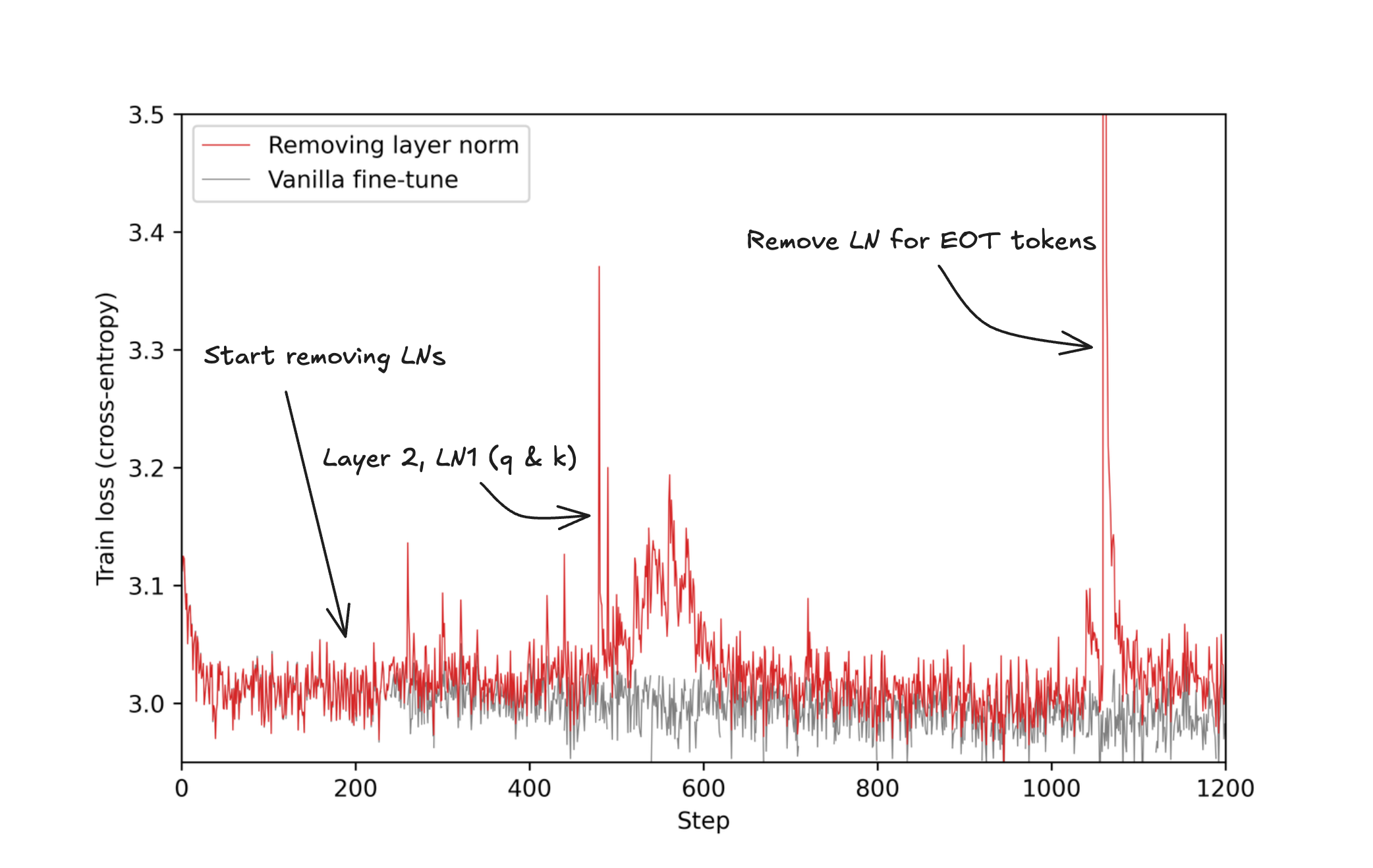}
  \caption{Removing LayerNorm while fine-tuning. The loss curve of a GPT2-small model
  being fine-tuned while gradually removing LN layers (red),
  compared to the loss from fine-tuning a vanilla GPT2-small model (gray).}
  \label{fig:main_loss_curves}
\end{figure}

\clearpage
\section{Introduction}
\label{sec:introduction}

Mechanistic interpretability aims to understand the inner workings of neural networks by
analyzing individual network components and their interactions (circuits). Recent work based on
sparse dictionary learning made progress in understanding the residual stream
\citep{Sharkey_Braun_Millidge_2022,cunningham2023sparse, bricken2023monosemanticity,
templeton2024scaling}, the attention layers \citep{kissane2024sparse,wynroe2024decomposing} and the
feed-forward layers
\cite{dunefskey2024transcoders}. Attribution techniques such as direct logic attribution
\citep{nostalgebraist2020interpreting,elhage2021mathematical,wang2022interpretability}, integrated
gradients \citep{friedman2004,sundararajan2017axiomatic,apolloIG}, and activation- and attribution
patching
\citep[e.g.][]{vig2020causal,geiger2021causal,meng2023locating,nanda2023attribution} have been used to
understand which model internals are responsible for the model's behavior.

All frontier LLMs are transformer models
\citep{brown2020language,touvron2023llama,openai2024gpt4,geminiteam2024family}.
Transformers consist of a residual stream, and a series
of components (attention layers, feed-forward layers) that read and write from the residual stream.
Of particular interest for this paper are the normalization layers that normalize the residual stream
as it is read by the attention and feed-forward layers, and a final normalization layer that
normalizes the residual stream before the unembedding.
These normalization layers are introduced to stabilize and speed up training of models \citep[as a
replacement for batch normalization,][]{ioffe2015batch} and are active at inference time (unlike batch
normalization layers). The two common choices are LayerNorm \citep[LN,][]{ba2016layer} or RMSNorm \citep{zhang2019root}. Both
operate on the embedding dimension of the residual stream.
\begin{gather}
  \text{LN}(\mathbf{x}) = \frac{\mathbf{x} - {\mu}}{\sigma} \odot \bm{\gamma} + \bm\beta
  \quad\quad\quad
  \text{RMSNorm}(\mathbf{x}) = \frac{\mathbf{x}}{\sigma} \odot \bm{\gamma}
  \\
  \text{where}\quad
  \mu = \frac{1}{H} \sum_{h=1}^{H} x_h
  \quad\quad\quad
  \sigma = \sqrt{\frac{1}{H} \sum_{h=1}^{H} (x_h - \mu)^2}
\end{gather}
At inference time, the mean centering ($\mu$), weight ($\bm\gamma$), and bias ($\bm\beta$)
parameters can be folded into neighboring layers \footnote{See e.g.
\href{https://github.com/TransformerLensOrg/TransformerLens/blob/main/further_comments.md\#what-is-layernorm-folding-fold_ln}{\texttt{fold\_ln}}
in TransformerLens \citep{nanda2022transformerlens}.},
so both normalizations are equivalent and
can be simplified as a division by the standard deviation\footnote{Note that the \enquote{standard deviation}
here is simply applied to an individual embedding vector, separately for each batch or token index.}
($\sigma$) of the embedding vector. For
simplicity, we will refer to both normalization layers as LN in the following.

These LN layers have been a hindrance to mechanistic interpretability over the last years. The
reasons mostly\footnote{The extra LN layers introduced in \citet{elhage2022solu} are introduced for
a different reason, and have and unrelated though also-hindering effect on interpretability.} fall
into three categories:
\begin{enumerate}
  \item Residual stream directions can not be directly interpreted as changes
  to the logits due to the final LN layer. This hinders logit lens analysis
  \citep[also known as direct logit
  attribution,][]{nostalgebraist2020interpreting,elhage2021mathematical,wang2022interpretability},
  as well as attribution patching \citep{nanda2023attribution}. \citet{olah2023mayupdate}
  refer to this as \enquote{reason \#78 for why interpretability researchers hate LayerNorm}.
  \item The transformer cannot be decomposed well into individual paths (circuits) without
  approximating LN layers.
  \citet{elhage2021mathematical} and
  \citet{sharkey2023technical} have argued that
  decomposing transformer models into individual
  circuits would be much easier without LN.
  In practice, \citet{bricken2023monosemanticity}, \citet{rushing2023copy}, and
  \citet{kissane2024sparse} all approximate (linearize) LN layers by freezing the normalization
  scale.
  \item We do not know whether the LN layers play an important role in the model's computation.
  Recent work on toy models \citep{winsor2022reexamining} showed that LN can be used as the sole
  non-linearity, \citep{stolfo2024confidence} suggest that LN might be used to implement confidence
  regularization in LLMs.
\end{enumerate}
In brief, it is a common occurrence to hear the phrase \enquote{turns out that LayerNorm completely
breaks things} \citep{nanda2023attribution} among mechanistic interpretability researchers.

In this paper, we demonstrate that, in GPT2-small, the LN layers can be removed after pre-training
by fine-tuning on a small fraction of the training data. Our primary goal is to show that an LLM of
near-identical capability to GPT2-small can be achieved without any LN layers. We propose that such
a model should be used as model organism for interpretability research, the role that is currently
played by the original GPT2-small model.
Previously, the only available language transformer models without LN were tiny models, such
as the 4-layer TinyModel \citep{noanabeshima2024tinymodel}.

We provide details of our fine-tuning procedure in Section \ref{sec:methodology}, present
loss-curves and final model benchmarks in Section \ref{sec:results}, and discuss applications
and open questions in Section \ref{sec:discussion}. We provide the fine-tuned GPT2-small model
in \href{https://huggingface.co/apollo-research/gpt2_noLN}{this} Hugging Face repository, including code to load the model
into the TransformerLens \citep{nanda2022transformerlens} 
library.

\section{Methodology}
\label{sec:methodology}

\begin{table}[b!]
  \caption{Training step at which we disable each LayerNorm layer. The digit before the dot
  indicates the transformer block, the name refers to the LN before the attention layer
  (\texttt{ln1qk} or \texttt{ln1v}), the feed-forward layer (\texttt{ln2}), or the unembedding
  (\texttt{lnf}). \texttt{eot} indicates the special case for the EOT tokens, and \texttt{bos}
  the special case for the first token.}
  \label{tab:run_details}
  \centering
  \small{\begin{tabular}{lrrrrr|lrrrrr}
    \toprule
    \textbf{Layer} & \textbf{v1} & \textbf{v2} & \textbf{v3} & \textbf{v4} & \textbf{v5} & \textbf{Layer} & \textbf{v1} & \textbf{v2} & \textbf{v3} & \textbf{v4} & \textbf{v5} \\
    \midrule
    \texttt{ 0.ln2}    & 50 & 50 & 200 & 200 & 180 &      \texttt{ 7.ln1v} & 50 & 350 & 1230 & 890 & 610 \\
    \texttt{ 1.ln2}    & 50 & 50 & 240 & 220 & 190 &      \texttt{ 8.ln1v} & 50 & 350 & 1240 & 920 & 620 \\
    \texttt{ 2.ln2}    & 50 & 50 & 280 & 240 & 200 &      \texttt{ 9.ln1v} & 50 & 350 & 1250 & 950 & 630 \\
    \texttt{ 3.ln2}    & 50 & 50 & 320 & 260 & 210 &      \texttt{10.ln1v} & 50 & 350 & 1260 & 980 & 640 \\
    \texttt{ 4.ln2}    & 50 & 50 & 360 & 280 & 220 &      \texttt{11.ln1v} & 50 & 350 & 1270 & 1010 & 650 \\
    \texttt{ 5.ln2}    & 50 & 50 & 400 & 300 & 230 &      \texttt{   lnf}  & 300 & 400 & 1640 & 1040 & 660 \\
    \texttt{ 6.ln2}    & 50 & 50 & 440 & 320 & 240 &      \texttt{ 0.eot}  & 200 & 500 & 1740 & 1060 & 680 \\
    \texttt{ 7.ln2}    & 50 & 50 & 480 & 340 & 250 &      \texttt{ 1.eot}  & 200 & 500 & 1740 & 1060 & 700 \\
    \texttt{ 8.ln2}    & 50 & 50 & 520 & 360 & 260 &      \texttt{ 2.eot}  & 200 & 500 & 1740 & 1060 & 720 \\
    \texttt{ 9.ln2}    & 50 & 50 & 560 & 380 & 270 &      \texttt{ 3.eot}  & 200 & 500 & 1740 & 1060 & 740 \\
    \texttt{10.ln2}    & 50 & 50 & 600 & 400 & 280 &      \texttt{ 4.eot}  & 200 & 500 & 1740 & 1060 & 760 \\
    \texttt{11.ln2}    & 50 & 50 & 640 & 420 & 290 &      \texttt{ 5.eot}  & 200 & 500 & 1740 & 1060 & 780 \\
    \texttt{ 0.ln1qk}  & 50 & 100 & 680 & 440 & 300 &      \texttt{ 6.eot}  & 200 & 500 & 1740 & 1060 & 800 \\
    \texttt{ 1.ln1qk}  & 50 & 120 & 720 & 460 & 320 &      \texttt{ 7.eot}  & 200 & 500 & 1740 & 1060 & 820 \\
    \texttt{ 2.ln1qk}  & 50 & 140 & 760 & 480 & 340 &      \texttt{ 8.eot}  & 200 & 500 & 1740 & 1060 & 840 \\
    \texttt{ 3.ln1qk}  & 50 & 160 & 800 & 500 & 360 &      \texttt{ 9.eot}  & 200 & 500 & 1740 & 1060 & 860 \\
    \texttt{ 4.ln1qk}  & 50 & 180 & 840 & 520 & 380 &      \texttt{10.eot}  & 200 & 500 & 1740 & 1060 & 880 \\
    \texttt{ 5.ln1qk}  & 50 & 200 & 880 & 540 & 400 &      \texttt{11.eot}  & 200 & 500 & 1740 & 1060 & 900 \\
    \texttt{ 6.ln1qk}  & 50 & 220 & 920 & 560 & 420 &      \texttt{ 0.bos}  & 200 & 700 & 2040 & 1160 & 920 \\
    \texttt{ 7.ln1qk}  & 50 & 240 & 960 & 580 & 440 &      \texttt{ 1.bos}  & 200 & 700 & 2040 & 1160 & 925 \\
    \texttt{ 8.ln1qk}  & 50 & 260 & 1000 & 600 & 460 &      \texttt{ 2.bos}  & 200 & 700 & 2040 & 1160 & 930 \\
    \texttt{ 9.ln1qk}  & 50 & 280 & 1040 & 620 & 480 &      \texttt{ 3.bos}  & 200 & 700 & 2040 & 1160 & 935 \\
    \texttt{10.ln1qk}  & 50 & 300 & 1080 & 640 & 500 &      \texttt{ 4.bos}  & 200 & 700 & 2040 & 1160 & 940 \\
    \texttt{11.ln1qk}  & 50 & 320 & 1120 & 660 & 520 &      \texttt{ 5.bos}  & 200 & 700 & 2040 & 1160 & 945 \\
    \texttt{0.ln1v}    & 50 & 350 & 1160 & 680 & 540 & \texttt{ 6.bos}  & 200 & 700 & 2040 & 1160 & 950 \\
    \texttt{1.ln1v}    & 50 & 350 & 1170 & 710 & 550 & \texttt{ 7.bos}  & 200 & 700 & 2040 & 1160 & 955 \\
    \texttt{2.ln1v}    & 50 & 350 & 1180 & 740 & 560 & \texttt{ 8.bos}  & 200 & 700 & 2040 & 1160 & 960 \\
    \texttt{3.ln1v}    & 50 & 350 & 1190 & 770 & 570 & \texttt{ 9.bos}  & 200 & 700 & 2040 & 1160 & 965 \\
    \texttt{4.ln1v}    & 50 & 350 & 1200 & 800 & 580 & \texttt{10.bos}  & 200 & 700 & 2040 & 1160 & 970 \\
    \texttt{5.ln1v}    & 50 & 350 & 1210 & 830 & 590 & \texttt{11.bos}  & 200 & 700 & 2040 & 1160 & 975 \\
    \texttt{6.ln1v}    & 50 & 350 & 1220 & 860 & 600 & lr-sched. & const & const & const & var & var \\
  \bottomrule
  \end{tabular}}
\end{table}

Previous works \citep[e.g.][]{heimersheim2023residual} observed the residual stream standard deviation
$\sigma$ does not vary a lot between different forward passes (except for end-of-text (EOT) tokens
used to indicate the beginning or end of sequences, and the first token in a prompt).
This suggests that replacing the per-token standard deviation $\sigma$ with a constant value $\bar\sigma$
calculated by averaging over a couple of prompts \citep[``freezing'' the normalization scale,
as done in][]{bricken2023monosemanticity,rushing2023copy,kissane2024sparse} may be possible.
We find that freezing all LN layers simultaneously breaks the model irreparably, resulting in either cross-entropy loss reaching NaN or remaining permanently $\gg 20$. However, a more gradual approach—freezing (parts of) the LN layers incrementally—yields a recoverable state. In this case, the loss initially increases, sometimes spiking to $\sim 20$, but the model can be fine-tuned to restore the loss to approximately its original value.

Our fine-tuning procedure contains three key ingredients:
\begin{enumerate}
  \item Disable one LN at a time. There are two LN layers in each transformer block,
  \texttt{ln1} before attention layer, and \texttt{ln2} before the feed-forward layer, plus the
  final layer norm \texttt{lnf}. We disable one LN layer in one block at a time, and fine-tune the
  model for a small number of steps.
  \item Treat \texttt{ln1} before the query and key vectors (\enquote{\texttt{ln1qk}}) separately
  from the \texttt{ln1} before the value vectors (\enquote{\texttt{ln1v}}). We noticed that the
  latter appeared to be more sensitive to freezing the LN scale, and disabling \texttt{ln1v} after
  \texttt{ln1qk} led to a more stable fine-tuning procedure.
  \item Handle the first sequence position, and EOT tokens, separately. In these situation the
  standard deviation tends to be much larger \citep{heimersheim2023residual}, so we use a second
  fixed $\bar\sigma_{0}$ value for these cases. We use the special case for the sequence
  position for all LN layers, but the special case for EOT tokens was only necessary for
  \texttt{ln1v}. Towards the end of the fine-tuning procedure, we remove these
  special cases one by one.
\end{enumerate}

We collect the average standard deviations from 16 OpenWebText prompts, using the first token to calculate
$\bar\sigma_{0}$, and the remaining tokens to calculate $\bar\sigma$. We then fine-tune GPT2-small
on the OpenWebText dataset \citep{Gokaslan2019OpenWeb}.
We use a batch size of 48, with 10
gradient accumulation steps, and a sequence length of 1024. We refer to 10 batches (i.e. one full
gradient accumulation) as one step,
containing 491,520 tokens. We use a base learning rate of $6\cdot10^{-4}$, and optionally use a
linear learning rate warm-up for 100 steps and cosine decay schedule to decrease
the learning rate to $6\cdot 10^{-5}$ after 2000 steps. Most models are trained for around 1000
steps, which corresponds to around 500M tokens, and takes around 2 hours on a single A100 GPU.

We start with $\sim200$ steps with all LN layers enabled, then disable LN layers one by one,
and then disable the special case for the first token and EOT tokens (optionally for one layer at a time).
We run a sweep of experiments using different LN removal schedules (sometimes removing LNs layers in
multiple blocks at a time) and report the best-performing schedules in Table \ref{tab:run_details}.

We evaluate the final LN-free models on the OpenWebText dataset, the ThePile
\citep[][via \texttt{apollo-research/monology-pile-uncopyrighted-tokenizer-gpt2}]{gao2020thepile} dataset (cross-entropy loss), and the Hellaswag
\citep{zellers2019hellaswag} benchmark \citep[using the implementation by][]{karpathy2022nanogpt}.
To provide a fair comparison on the OpenWebText dataset---as our model was fine-tuned on
(a different section of) this dataset---we also fine-tune a \enquote{vanilla} GPT2-small model
(with all LN layers enabled) for 1000, 1200, and 2000 steps, and report its performance.

\section{Results}
\label{sec:results}
\renewcommand*{\thefootnote}{\fnsymbol{footnote}}
\begin{table}[b]
  \caption{Comparison of model performance between the fine-tuned no-LN model, the original
  GPT2-small model,
  and equivalently fine-tuned vanilla models.
  Note: Runs v4 and v5 used the variable learning rate schedule, so for comparison
  these should be compared to the starred vanilla model (which uses
  the same schedule).}
  \label{tab:model_comparison}
  \centering
  \begin{tabular}{lccc}
    \toprule
    \textbf{Version} & \textbf{OpenWebText} & \textbf{ThePile} & \textbf{Hellaswag} \\
    \midrule
    no-LN v1 (850 steps) & 3.130 & 3.057 & 29.17\% \\ %
    no-LN v2 (1000 steps) & 3.014 & 2.926 & 29.54\% \\ %
    no-LN v3 (2450 steps) & 3.010 & 2.931 & 29.39\% \\ %
    no-LN v4 (1200 steps\footnotemark[1]) & \textbf{3.000} & \textbf{2.900} & 29.51\% \\ %
    no-LN v5 (1100 steps\footnotemark[1]) & 3.006 & 2.930 & 29.52\% \\ %
    \midrule
    Original (no fine-tuning) & 3.095 & 2.856 & 29.56\% \\
    Vanilla (1000 steps) & 2.989 & 2.880 & 29.82\% \\
    Vanilla (2000 steps) & 2.978 & 2.905 & 29.64\% \\
    Vanilla (1200 steps\footnotemark[1]) & \textbf{2.966} & \textbf{2.850} & 30.01\% \\
    \bottomrule
  \end{tabular}
\end{table}
\renewcommand*{\thefootnote}{\arabic{footnote}}

We are able to successfully train a GPT2-small model without any LN layers (\enquote{no-LN} model).
Our best model (v4) achieves a cross-entropy loss of 3.000 on the OpenWebText dataset
(compared to 2.966 for the fine-tuned vanilla GPT2-small model with LN). On ThePile our same model
achieves a loss of 2.9000 (compared to 2.850 for the baseline with LN).
Additionally we report the accuracy on the Hellaswag benchmark, which is 29.5\% (compared
to 30.0\% for the original GPT2-small model).
Table \ref{tab:model_comparison} shows the performance of the 5 different fine-tuned models (v1-v5)
as described in Table \ref{tab:run_details}.
We also provide generations from a no-LN model, compared to the vanilla model, in Appendix
\ref{sec:appendix}. There is no notable difference in the quality of the generations.

To put this loss difference into perspective, we estimate what model size would be needed to achieve
this loss at the same compute budget. Following the Chinchilla-based scaling laws
\citep{hoffmann2024training} we roughly estimate \citep[using the graphs from][]{korbak2022training}
that a 0.05 cross-entropy loss difference corresponds to a 93M (rather than 117M) parameter model.

\begin{figure}[h!]
  \centering
  \includegraphics[width=\textwidth]{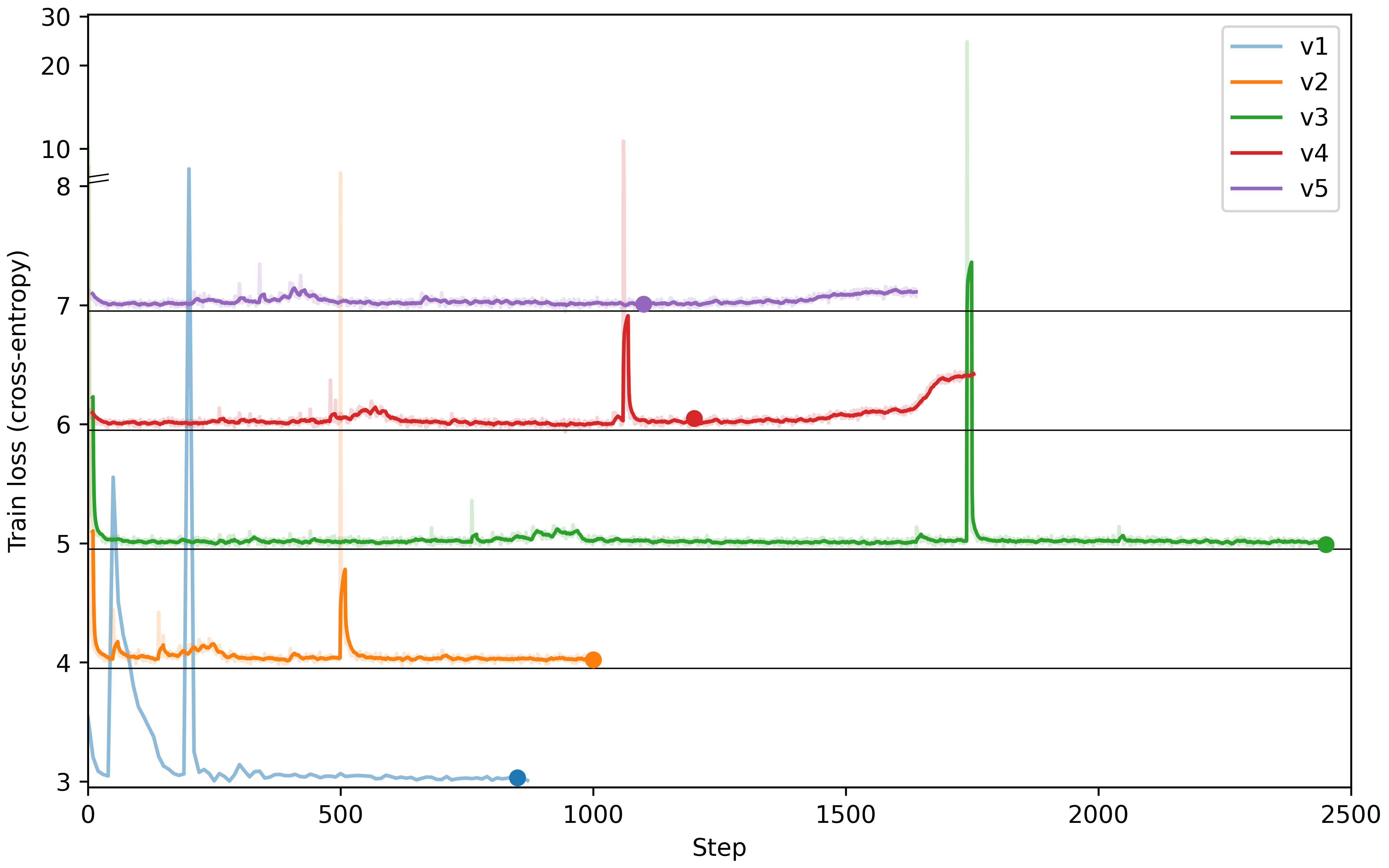}
  \caption{Training loss curves for all 5 no-LN models. The dot indicates
  the snapshot of best validation loss, which is the one we use for the benchmarks.
  Note that the y-axis is offset by 1 for each version,
  and log-scaled for $y>8$.}
  \label{fig:loss_curves}
\end{figure}

Figure \ref{fig:loss_curves} shows the training loss curves for all 5 no-LN models, and Figure
\ref{fig:main_loss_curves} shows the loss curve for the best no-LN model (v4) compared to
equivalently fine-tuning a vanilla GPT2-small model.
We observe jumps in the loss curve when disabling the LN layers, as expected, but in most cases
(the ones shown here) the model is able to recover from these jumps. Empirically we found that
disabling many LN layers at once, or in quick succession, leads to a significant loss spike from
which the model does not fully recover. For example, in the no-LN v1 we disabled many LN layers
at once, and the OpenWebText validation loss never drops below 3.1 after that.

The last run, no-LN v5, is a run where we spread out the LN removal as widely as possible
(every removal happens at a different step). We find that we can indeed avoid large loss spikes,
although the model does not end up being our best-performing model (possibly due to still removing
LN layers in quick succession).

\section{Discussion}
\label{sec:discussion}
We want to discuss the application of this technique, and the loss penalty for removing
LayerNorm. The cross-entropy loss drop of 0.05 is a significant drop in terms of modern LLMs,
and we would not expect a production model to remove LN layers. However, our goal is not primarily
to make SOTA model interpretable, but to work towards a full mechanistic understand of \textit{any}
large language model.
We don't know whether mechanistic interpretability insights from the no-LN version of a model
would transfer to the original model, i.e. whether the no-LN model is sufficiently faithful to
the original model. However, the transfer of other techniques \citep[e.g.][]{kissane2024saestransfer}
suggests that they might. This would allow us to reach a secondary goal of (eventually)
understanding SOTA LLMs.

The high performance of the no-LN models suggest that the LN layers do not play an important role
in language modelling. This provides evidence that the common practice of linearizing LN
\citep{bricken2023monosemanticity,rushing2023copy,kissane2024sparse} probably does not obscure
important model behaviour.\footnote{To be clear, the model (without fine-tuning) absolutely does not
work when LN is linearized, but the fact that it can adjust to this change (with fine-tuning)
relatively easily is a good sign.}

In this work we only consider the GPT2-small model, which
leads to two limitations: (1) It is possible that LN layers play a more important role in larger
models and it is not possible to remove LN there. (2) Training larger models is harder, so this
fine-tuning procedure might be more difficult or more expensive for larger models.

We also want to highlight some confusing aspects of our results: (1) We found that the loss on
ThePile drops more than on OpenWebText when removing LN, suggesting perhaps a worse generalization.
However this also brings the losses on ThePile and OpenWebText closer together, so we are unsure about
the conclusion on generalization. (2) We noticed that in the runs with variable learning rate schedule (v4 and v5), the
the loss curves started to rise towards the end of training (long after removing the LNs). We do
not understand why this happens when the learning rate is decreased.

There are a few improvements that we would like to see in future works. First, it might be helpful
to collect more data to compute the averages $\bar\sigma$ and $\bar\sigma_0$, and possibly to
separate $\bar\sigma_0$ into separate averages for position 0 tokens and EOT tokens. Second, we
only briefly explored the idea of gradually turning off the individual LN layers (not shown in this paper).\footnote{This
can be done by using a combination of $\sigma$ and $\bar\sigma$ in the LN formula, and slowly
increasing the weight of $\bar\sigma$ while fine-tuning.} Such a gradual removal might help reduce the loss spikes
further. Finally, and most importantly, we would like to see this technique applied to much larger
models.

\section{Conclusion}
This paper demonstrates that the LayerNorm layers in GPT2-small can be removed after pre-training
by fine-tuning on a small fraction of the training data (500M tokens, 2 GPU-hours). We make our
trained models available in a Hugging Face repository, and the models are already being used in
mechanistic interpretability research \citep[work in progress,][]{giglemiani2023personal,janiak2023personal}

The removal of LayerNorm allows researchers to leverage the success of sparse dictionary learning in
understanding individual components of the transformer model \citep{sharkey2024sparsify,
templeton2024scaling,wynroe2024decomposing,dunefskey2024transcoders} and analyze the interaction
between these components, putting it all together.

While frontier models will likely continue to be trained with LN (or a similar normalization layer),
we believe that understanding any capable LLM (e.g. a GPT2-small model) would be a major success of
mechanistic interpretability. Techniques or insights transferring to larger models with LN would be
a bonus.

\subsubsection*{Acknowledgments}
Thanks to Alice Rigg, Bilal Chughtai, Leo Gao, Neel Nanda, and Rudolf Laine for comments and feedback on the draft and idea. The
\href{https://github.com/karpathy/nanoGPT}{nanoGPT} repository and accompanying
\href{https://www.youtube.com/watch?v=l8pRSuU81PU}{video} by Andrej Karpathy helped me implement
this idea quickly.

{\small \bibliography{references}}

\appendix
\section{Appendix}
\label{sec:appendix}
We provide some samples of text generated by the no-LN GPT2-small model, and the vanilla GPT2-small
model.

\subsection{GPT2-small without LayerNorm (v2 model)}
Two example generated texts from the no-LN GPT2-small model (v2) are shown below.

\begin{displayquote}
  As the last leaf fell from the tree, John realized that he was going to have to use it. “We’ve all been there. We’ve all been there,” he said. “It’s been a long time. But, it’s good to be back. It’s good to be back.” “It’s good to be back. It’s good to be back
\end{displayquote}

\begin{displayquote}
  As the last leaf fell from the tree, John realized that a large amount of weight had been lifted from him. "I had a little panic attack. I was afraid that I could not walk," he said. "I felt like my legs were going to break." John has since gone back to the tree. "I have to tell you that I'm sorry I did that, but I don't know if that will ever happen," he said.
\end{displayquote}

\subsection{Vanilla GPT2-small model}
Two example generated texts from the vanilla GPT2-small model are shown below.

\begin{displayquote}
  As the last leaf fell from the tree, John realized that it was empty. He took the leaf and turned it over to his wife, who told him that it was still there and that he would have to go to the church to find it. John went to the church, and found that it was empty. He said, "I am going to the church and I am going to find the rest of the leaves, and I am going to look for them and find out where they
\end{displayquote}

\begin{displayquote}
  As the last leaf fell from the tree, John realized that the tree had been torn down. As he turned his head, the other trees started to fall. "Come on," John said, "we're going to get out of here!" The next tree was a wildflower. "How is it?" John asked, "do you see any other way?" "It's a good thing," the other trees replied.
\end{displayquote}

\end{document}